\documentclass[letterpaper, 10 pt, conference]{ieeeconf} 
\IEEEoverridecommandlockouts    

\usepackage{glossaries}

\newacronym{mav}{UAV}{Unmanned Aerial Vehicle}
\newacronym{uav}{UAV}{Unmanned Aerial Vehicle}
\newacronym{ovc}{OVC}{Open Vision Computer}
\newacronym{lidar}{LiDAR}{Light Detection and Ranging}
\newacronym{vio}{VIO}{visual-inertial odometry}
\newacronym{satnav}{GNSS}{Global Navigation Satellite Systems}
\newacronym{dgps}{DGPS}{Differential GPS}
\newacronym{rtk}{RTK}{Real-Time Kinematics}
\newacronym{ppk}{PPK}{Post-Processed Kinematic}
\newacronym{gpgpu}{GPGPU}{General-Purpose Graphics Processing Unit}
\newacronym{hri}{HRI}{Human-Robot Interaction}
\newacronym{ugv}{UGV}{Unmanned Ground Vehicle}
\newacronym{uwb}{UWB}{Ultra Wideband}
\newacronym{svm}{SVM}{Support Vector Machine}
\newacronym{fcn}{FCN}{Fully Convolutional Network}
\newacronym{cnn}{CNN}{Convolutional Neural Network}
\newacronym{loam}{LOAM}{LiDAR Odometry and Mapping}
\newacronym{sloam}{SLOAM}{Semantic LiDAR Odometry and Mapping}
\newacronym{slam}{SLAM}{Simultaneous Localization and Mapping}
\newacronym{iot4ag}{IoT4Ag}{NSF Engineering Research Center for the Internet of Things for Precision Agriculture}
\newacronym{grasp-lab}{GRASP Lab}{the General Robotics, Automation, Sensing and Perception Laboratory}
\newacronym{jps}{JPS}{Jump Point Search}
\newacronym{ukf}{UKF}{Unscented Kalman Filter}
\newacronym{sam}{SAM}{Smoothing and Mapping}
\newacronym{icp}{ICP}{Iterative Closest Point}
\newacronym{imu}{IMU}{Inertial Measurement Unit}
\newacronym{tsdf}{TSDF}{Truncated Signed Distance Field}
\newacronym{esdf}{ESDF}{Euclidean Signed Distance Field}
\newacronym{rrt}{RRT}{A rapidly exploring random tree}

\let\labelindent\relax

\usepackage[table,usenames,dvipsnames]{xcolor}      
\usepackage{extarrows}                              
\usepackage{enumitem}
\usepackage{wrapfig}

\usepackage{amssymb,amsfonts,amsthm,dsfont} 
\usepackage{algorithm,algorithmicx,listings}        
\usepackage[noend]{algpseudocode}			              
\makeatletter
\def\BState{\State\hskip-\ALG@thistlm}
\makeatother

\usepackage{graphics} 
\usepackage{epsfig} 
\usepackage{times} 
\usepackage{amsmath} 

\usepackage{tabularx}
\usepackage{subcaption}
\usepackage{textcomp}

\usepackage[font={small}]{caption}   
\usepackage[breaklinks=true, colorlinks, bookmarks=true, citecolor=Black, urlcolor=Violet,linkcolor=Black]{hyperref}

\usepackage{svg}

\usepackage{amssymb,graphicx}

\usepackage{everypage}
\usepackage{afterpage}
\usepackage{ifthen}

\usepackage[normalem]{ulem}
\usepackage[utf8]{inputenc}
\usepackage[english]{babel}
\usepackage{hyperref}
\hypersetup{
    colorlinks=true,
    linkcolor=blue,
    filecolor=magenta,      
    urlcolor=cyan,
}

\DeclareMathAlphabet\mathbfcal{OMS}{cmsy}{b}{n}

\newtheorem*{assumption*}{Assumption}

\newtheorem*{problem*}{Problem}

\usepackage{lipsum}
\usepackage[capitalise]{cleveref}
\usepackage{svg}
\usepackage{mathtools}

\usepackage{tikz}
\newcommand\copyrighttext{%
  \footnotesize \textcopyright 2023 IEEE. Personal use of this material is permitted.
  Permission from IEEE must be obtained for all other uses, in any current or future
  media, including reprinting/republishing this material for advertising or promotional
  purposes, creating new collective works, for resale or redistribution to servers or
  lists, or reuse of any copyrighted component of this work in other works.}
\newcommand\copyrightnotice{%
\begin{tikzpicture}[remember picture,overlay]
\node[anchor=south,yshift=10pt] at (current page.south) {\fbox{\parbox{\dimexpr\textwidth-\fboxsep-\fboxrule\relax}{\copyrighttext}}};
\end{tikzpicture}%
}

\begin{document}

\title{\LARGE \bf Active Metric-Semantic Mapping by Multiple Aerial Robots}

\author{Xu Liu, Ankit Prabhu, Fernando Cladera, Ian D. Miller, Lifeng Zhou, Camillo J. Taylor, Vijay Kumar
\thanks{This work was supported by funding from the IoT4Ag ERC funded by the National Science Foundation (NSF) under NSF Cooperative Agreement Number EEC-1941529, NIFA grant 2022-67021-36856, NSF grant CCR-2112665, and C-BRIC, a Semiconductor Research Corporation Joint University Microelectronics Program cosponsored by DARPA. Ian D. Miller acknowledges the support of a NASA Space Technology Research Fellowship. We gratefully acknowledge Alex Zhou for the hardware support, and Guilherme V. Nardari for contributing to the factor graph implementation.}
\thanks{X. Liu, A. Prabhu, F. Cladera, I. D. Miller, C. J. Taylor, V. Kumar are with GRASP Laboratory, University of Pennsylvania {\tt\small\{liuxu, praankit, fclad, iandm, cjtaylor, kumar\}@seas.upenn.edu}.}
\thanks{L. Zhou was with GRASP Laboratory, University of Pennsylvania while completing this work. Presently, he is with the Department of Electrical and Computer Engineering, Drexel University \tt\small{lz457@drexel.edu.}}
\thanks{Digital Object Identifier (DOI): 10.1109/ICRA48891.2023.10161564.}
}

\maketitle
\copyrightnotice

\begin{abstract}
Traditional approaches for active mapping focus on building geometric maps. For most real-world applications, however, actionable information is related to semantically meaningful objects in the environment. We propose an approach to the active metric-semantic mapping problem that enables multiple heterogeneous robots to collaboratively build a map of the environment. The robots actively explore to minimize the uncertainties in both semantic (object classification) and geometric (object modeling) information. We represent the environment using informative but sparse object models, each consisting of a basic shape and a semantic class label, and characterize uncertainties empirically using a large amount of real-world data. Given a prior map, we use this model to select actions for each robot to minimize uncertainties. The performance of our algorithm is demonstrated through multi-robot experiments in diverse real-world environments. The proposed framework is applicable to a wide range of real-world problems, such as precision agriculture, infrastructure inspection, and asset mapping in factories.

\end{abstract}



\section{Introduction}
\label{sec:introduction}

\begin{figure}[t!]
        \centering
        \begin{subfigure}[t]{1.0\columnwidth}
        \centering       
        \includegraphics[trim=0 0 0 0, clip,width=\textwidth]{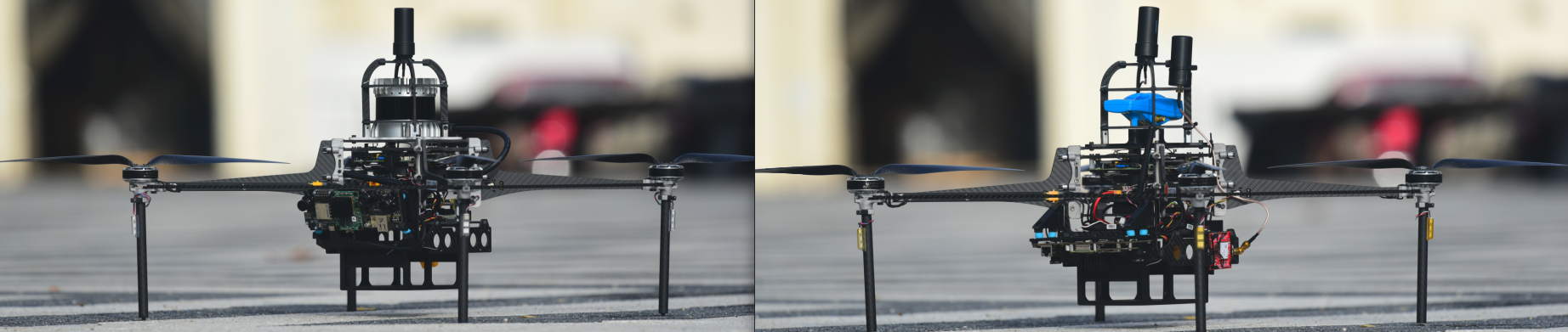}
        \vspace{-0.125in}
        \end{subfigure}
        \centering       
        \begin{subfigure}[t]{1.0\columnwidth}
        \centering
        \includegraphics[trim=0 175 0 150, clip,width=\textwidth]{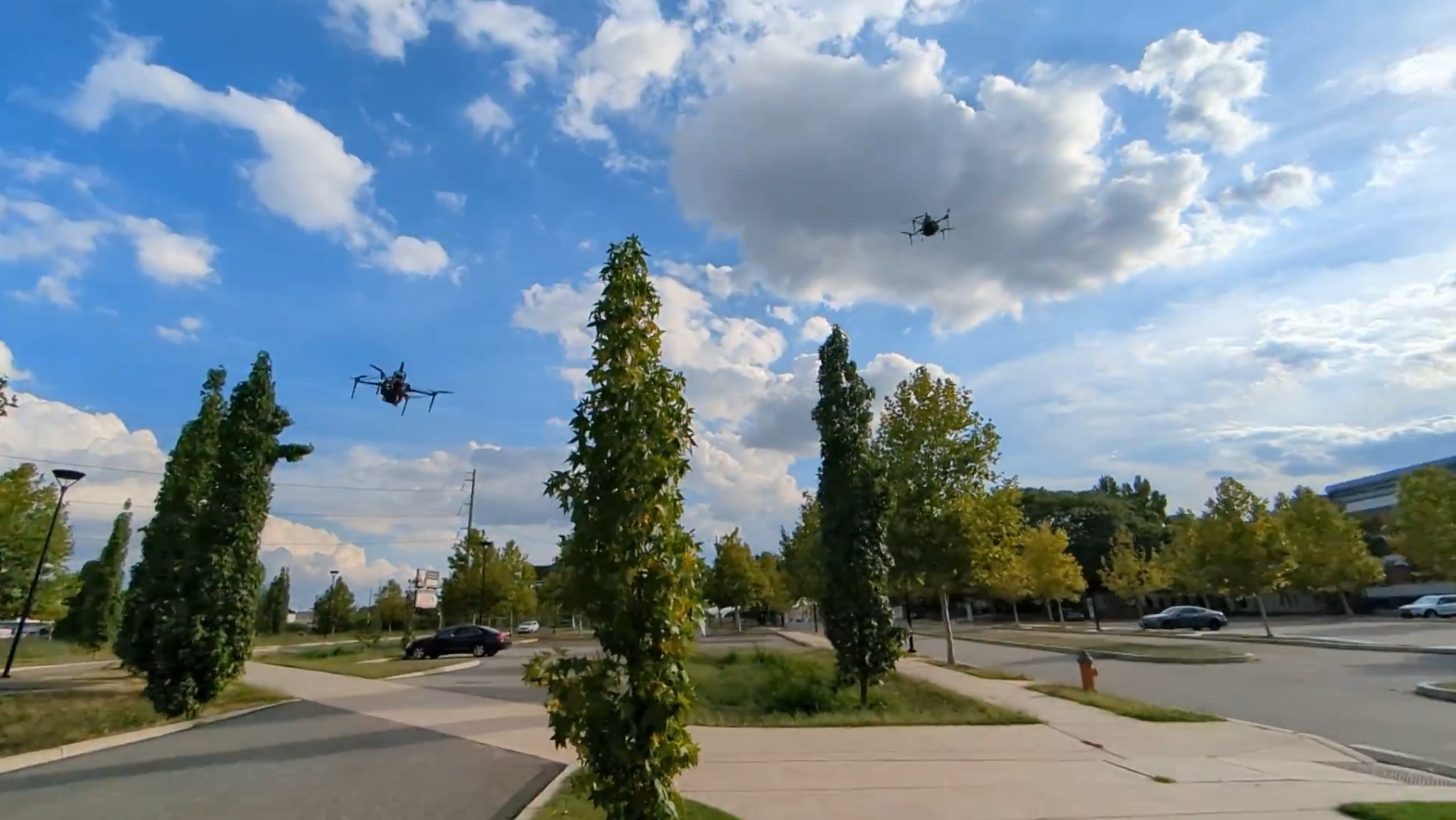}
        \end{subfigure}       
        \caption{(Top left) Low-altitude UAV for map refinement. (Top right) high-altitude UAV for aerial mapping. (Bottom) Active metric-semantic mapping by two autonomous UAVs without using GPS.}
        \label{fig:front-page}
        \vspace{-0.25in}
\end{figure}

Robots that can perceive and understand both semantic (e.g. class, species) and metric (e.g. shape, dimension) aspects of the environment and actively build metric-semantic maps can have a huge impact in many real-world applications. An example of our system is shown in \cref{fig:front-page}.

Modeling the environment using a set of semantically meaningful object models is important for long-range exploration and large-scale mapping tasks. Such a model stores actionable information about the environment critical for robot exploration. In addition, semantic maps can provide long-term localization constraints for robot teams (loop closure, intra-robot registration) due to their viewpoint invariance. Finally, sparse semantic representations can significantly reduce the storage requirements, and are suitable for multi-robot settings with limited robot-to-robot communication bandwidth. Due to these motivations and advancements in data-driven object detection, metric-semantic \gls{slam} methods are gradually becoming the new state-of-the-art in \gls{slam}.

\begin{figure*}[t!]
\centering
\includegraphics[trim=0 0 0 0, clip,width=1.0\textwidth]{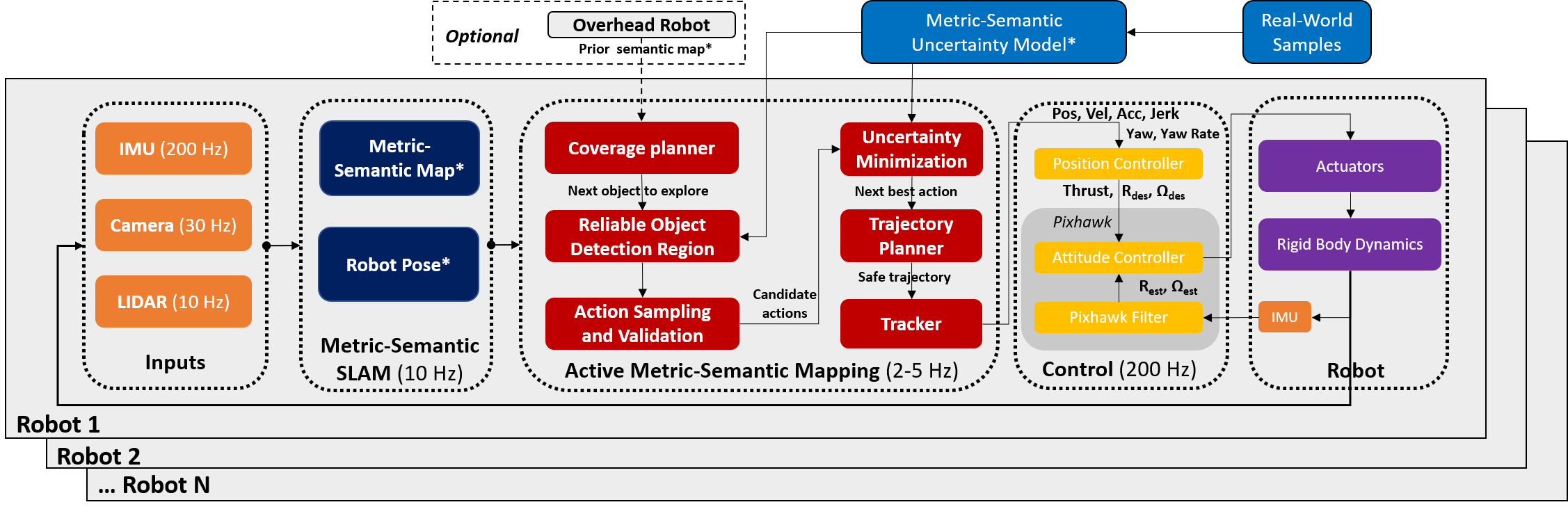}
    \vspace{-0.1in}
    \caption{{\textbf{System architecture}. The quantities shared between robots are marked by asterisks, including the metric-semantic map (object models and labels), robot poses, semantic and geometric uncertainty distributions, and the prior semantic map from an overhead robot. }}
        \vspace{-0.2in}
    \label{fig:factor-graph}
\end{figure*}

Despite the success of metric-semantic \gls{slam} methods in various environments~\cite{bowman2017probabilistic, nicholson2018quadricslam, atanasov2018unifying, yang2019cubeslam, rosinol2021kimera, chen2019suma++, chen2020sloam}, active metric-semantic mapping remains an open and challenging research problem~\cite{placed-atanasov-carlone-2022survey-active-slam}. This problem requires robots to infer changes not only in geometric but also in semantic uncertainties as a result of their actions. The measurement model is complex and varies with viewing angle, distance, occlusions, robot motion, and object surface properties. Also, the noise models for geometric maps are not valid for semantic objects. Furthermore, characterization of uncertainties for semantic object classification and modeling is a challenging task~\cite{rosen-Leonard-2021advances-survey}. 

In this paper, we propose a novel approach to the active metric-semantic mapping problem. Instead of making assumptions about the measurement model uncertainty or using heuristics for exploration, our algorithm empirically characterizes the noise from real-world observations. This can be intuitively thought of as allowing the robot to infer the uncertainty distributions based on its past observations of the world. The \textbf{contributions} of this paper include:

\begin{enumerate}
    \item We propose a real-time metric-semantic \gls{slam} algorithm that uses a generic, storage-efficient, semantically meaningful, and geometrically accurate environment representation, encodes object-robot constraints via customized factors in the factor graph to minimize robot odometry drift, and supports multi-robot collaboration. 
    \item We propose an active metric-semantic mapping algorithm built on the foundation of our metric-semantic \gls{slam} algorithm. This algorithm uses empirical uncertainty characterizations from real-world data.
    \item We integrate these algorithms with a complete autonomy stack and perform experiments in various real-world environments, including merging maps from multiple UAVs. The system is quantitatively demonstrated to gather higher quality information about objects of interest compared to benchmark methods. 
\end{enumerate}

To our knowledge, this work is the first to propose an active metric-semantic mapping system that enables the robot to minimize both metric and semantic uncertainties and is grounded in a systematic and empirical uncertainty characterization from a large amount of real-world data. A demo video of our system can be found at \url{https://youtu.be/S86SgXi54oU}.

\section{Related Work}
\label{sec:related-work}

We begin by discussing prior work in metric-semantic mapping, then active mapping, and finally works combining these areas into active metric-semantic mapping.

\subsection{Metric-semantic Mapping}
Many works in metric-semantic mapping represent the environment using dense semantically-annotated maps such as mesh~\cite{rosinol2021kimera}, surfel~\cite{chen2019suma++}, or 2.5D grid maps~\cite{maturana2018real}. These maps can be conveniently used for planning and navigation but have large demands on computation and storage. It is challenging to use these approaches for large-scale mapping and exploration due to limited onboard computation. These works also do not employ semantics to improve localization accuracy. Additionally, individual object models, often desirable for downstream tasks such as asset mapping, are not used in the \gls{slam} optimization. By contrast, other works build object-level maps using pre-collected 3D models~\cite{salas2013slam++}, or basic shapes such as points~\cite{bowman2017probabilistic}, cuboids~\cite{yang2019cubeslam}, ellipsoids~\cite{nicholson2018quadricslam}, cylinders~\cite{chen2020sloam}, a set of semantic keypoints~\cite{shan2020orcvio}, or a combination of 2D shapes~\cite{cao2021lidar}. Some efforts also account for ambiguity in data association~\cite{bowman2017probabilistic}, and the usage of semantic information in identifying moving objects~\cite{chen2021moving}, generating hierarchical descriptors for loop closure~\cite{hughes2022hydra}, and minimizing odometry drifts to assist long-range navigation~\cite{liu2022large}. 
However, these approaches do not define a measurement model that maps from state space to object classification confidence space, which is still an open and challenging problem \cite{rosen-Leonard-2021advances-survey}. Without such a model, these approaches cannot be adopted for active information acquisition in metric-semantic maps, where the robot needs to infer how semantic and geometric uncertainties are affected by its actions.

\subsection{Active mapping}
Prior works on active mapping mostly use geometric maps, which represent the environment using either sparse (e.g. landmark-based) or dense (e.g. volumetric) elements. \cite{atanasov2014information} proposes an efficient active information acquisition algorithm for sparse maps. \cite{schlotfeldt2018anytime} extends this work by decentralizing it for multiple robots, and adaptively adjusting the sub-optimality to satisfy the computation budget. For dense maps, \cite{charrow2015information1st} proposes an information-theoretic active 3D occupancy grid mapping algorithm heavily optimized to run in real time. To explore larger environments, some works employ a two-stage approach where the algorithm first plans paths for coverage and then refines the path for maximizing information gain or minimizing execution time~\cite{charrow2015information2nd-dense3dmapping, zhou2021fuel}. However, none of these methods account for uncertainties in semantic information, which we show often has a significantly different distribution than geometric uncertainties.

\subsection{Active Metric-semantic mapping}
Some prior works seek to characterize the uncertainty related to semantic objects~\cite{yu2019variational, cortinhal2020salsanext}, but cannot predict uncertainties of future measurements. Others address this next-best viewpoint prediction problem~\cite{atanasov2014nonmyopic, doumanoglou2016recovering} but verify the models only under controlled indoor conditions with prior knowledge, such as detailed 3D object models.
Finally, some works take a more end-to-end approach using reinforcement learning~\cite{chaplot2021seal} or map prediction~\cite{georgakis2021learning}, but validate their algorithms only in simulation.
Others take a more model-based approach using Gaussian Mixture Models~\cite{wang2019semantic} or Bayesian OcTrees~\cite{asgharivaskasi-atanasov-2021active-semantic-mapping, asgharivaskasi-atanasov-2022active-semantic-mapping-Octree} to model uncertainties. However, none of them explicitly model the relationship between object classification uncertainty and states of robot and object.

We approach the active metric-semantic mapping problem using model-based information-theoretic exploration, where the uncertainties of the metric-semantic measurement model are characterized empirically based on a large amount of real-world data. We leverage our semantic \gls{slam} module to automate this characterization process. Our approach allows robots to explore and build an uncertainty-minimized metric-semantic map in real time.


\section{Preliminaries and Problem Formulation}

\label{sec:problem-formulation}

\subsection{Preliminaries} 
Let there be $k$ robots $\{r^1, r^2, ..., r^k\}$.
The semantic map of the $k$th robot $\mathcal{M}^k$ consists of a set of semantic objects $\{\mathbf{\ell}^k_1, \ell^k_2, ..., \ell^k_{n}\}$ belonging to $v$ semantic classes $\{s_1, s_2, ..., s_v\} \in \mathcal{S}$.  For notational compactness, we suppress $k$ unless otherwise noted.  Each object has a state vector $\ell_i = \{\ell_i^s, \ell_i^g\}$ that defines the semantic and geometric properties of the object, where $\ell_i^s = (p_i, s_i) \in [0, 1] \times \mathcal{S}$, $p_i$ is the probability of $\ell_i$ belonging to class $s_i$, and $\ell_i^g$ is a vector defining the pose and geometric model of the object. The semantic class of the object defines the shape of the object, which is rectangular cuboid, cylinder, or plane\footnote{Planar objects are associated with ground classes and are implicitly used to constrain the bottom of the cylinder and cuboid models.}. For each cuboidal object, the state vector is (suppressing subscript $i$): $\ell^{g} = [\mathbf{r}; \mathbf{t}; \mathbf{d}]$, where $\mathbf{r}$ = [$r_x$,  $r_y$,  $r_z$]$^\intercal$ is the rotation vector, 
$\mathbf{t}$ = [$t_x$,  $t_y$,  $t_z$]$^\intercal$ is the translation vector, and 
$\mathbf{d}$ = [$d_x$,  $d_y$,  $d_z$]$^\intercal$ is the dimension vector. For each cylindrical object, the state vector is:
    $\ell^{g} = [\mathbf{b}; \mathbf{n}; r]$,
where $\mathbf{b}$ = [$b_x$,  $b_y$,  $b_z$]$^\intercal$ is the origin of the axis ray,
$\mathbf{n}$ = [$n_x$,  $n_y$,  $n_z$]$^\intercal$ is the direction of the axis ray, and 
$r$ is the radius. The robot state at time $t$ is represented by $\mathbf{x}_t$, which contains the $\mathbb{SE}$(3) pose.

We assume that the base (take-off) locations of robots are close and known relative to each other, and allow robots to communicate and exchange information only at the base. Note that this implies that the robots explore independently, but can collaboratively construct semantic maps when at the base. Also, simultaneous and sequential flights are equivalent under this assumption. In our prior work \cite{miller2022stronger}, we proposed a method for localizing robots using a semantic map built by a high-altitude UAV and opportunistically communicating data within the team. Using these methods, we can relax our assumptions and extend our work to handle intermittent communication or unknown take-off positions. 

\begin{figure}[t!]
    \centering
    \includegraphics[trim=0 0 0 10, clip, width=0.95\linewidth]{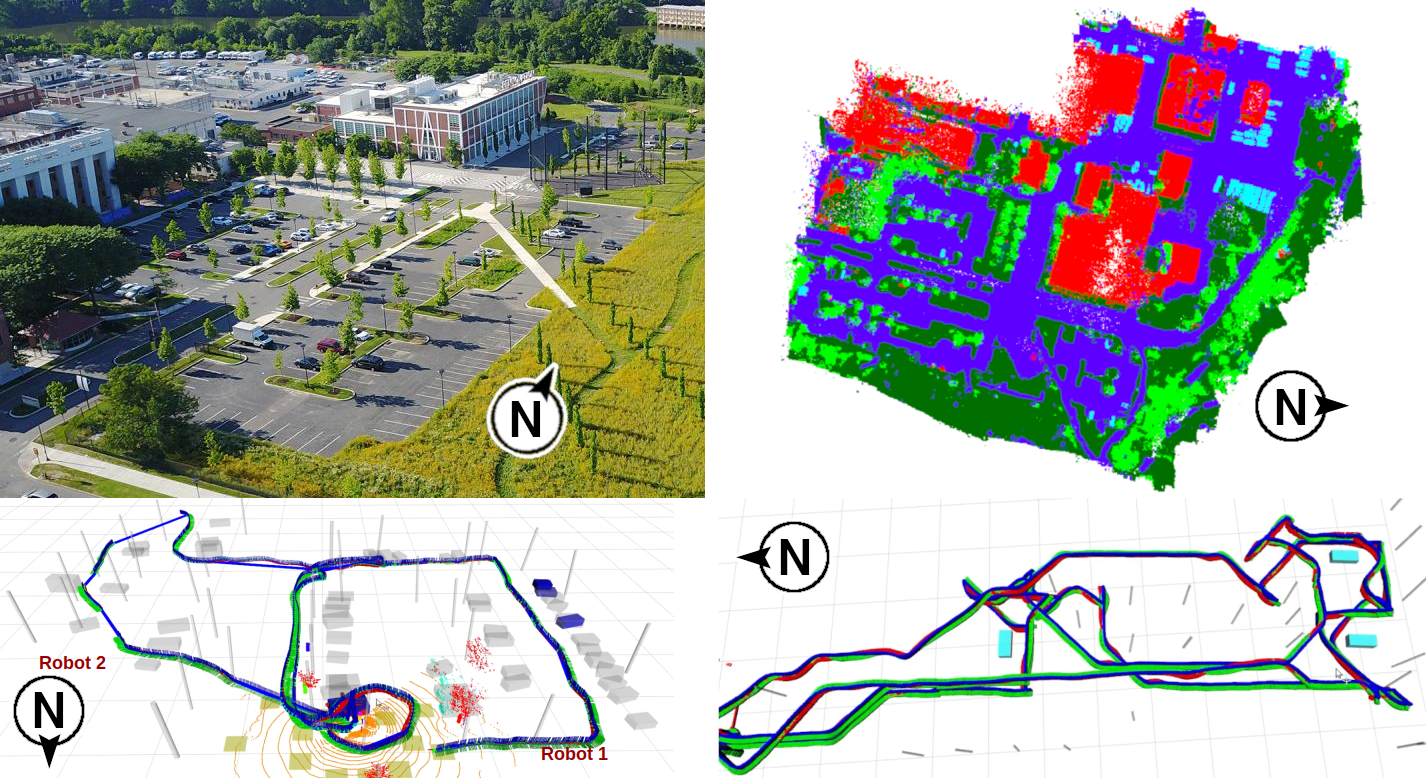}
    \caption{\textbf{Active semantic mapping of urban areas with a heterogeneous team of UAVs}. (Top left) Overhead view of one of our experiment environments. (Top right) Semantic map built by the overhead robot in real time, where vehicles are cyan colored. (Bottom) Examples of the semantic map built by two low-altitude robots. Cylinders represent light poles or tree trunks, cuboids represent vehicles, and planes represent the local ground.}
    \label{fig:semantic maps}
\end{figure}

\subsection{Objective}
Given an unknown environment, or limited prior knowledge about the environment, our objective is to find $\hat{\mathcal{M}}$, such that $\|\hat{\mathcal{M}} \ominus {\mathcal{M}}\|$ is minimized. $\ominus$ is the generalized difference between the two maps. In other words, we want our estimated metric-semantic map to best approximate the real-world map. We decompose our problem into metric-semantic \gls{slam} and Planning for Active Metric and Semantic information acquisition (PAMS). 

\subsubsection{Metric-Semantic SLAM}
The objective of metric-semantic \gls{slam} is to accumulate previous measurements $\{\mathbf{z}_1, \mathbf{z}_2, ..., \mathbf{z}_t\} \in \mathcal{Z}_t$ to estimate the current metric-semantic map $\mathcal{M}_t$ and a set of robot trajectories \{$\mathcal{{T}}_1$, $\mathcal{{T}}_2$, .., $\mathcal{{T}}_k$\}.

\subsubsection{PAMS}
The objective of PAMS is to find the best trajectory for each robot that minimizes both \textbf{semantic} and \textbf{geometric} uncertainties, given $\mathcal{Z}_t$ and the initial states of all robots $\mathbf{x}_0$, and a planning horizon $\boldsymbol{\tau} \triangleq t+1 : t+T$. Recall that robots explore independently, so we treat PAMS as a single-robot problem. Minimizing semantic and geometric uncertainty requires confidently classifying an object, and reducing errors in geometric models of objects $\ell^g_i$. Instead of directly optimizing the two objectives together, we cast the classification confidences $p_i$ as the constraint of the optimization problem, and minimizing the uncertainties of $\ell^g_i$ as the objective\footnote{We assume that the robot pose estimation error is negligible, since we use an accurate LIDAR-inertial odometry algorithm to estimate frame-to-frame relative transformation and semantic landmarks to minimize the drift.}. 
In other words, we try to improve the accuracy of the object model, once the object exists and belongs to the class of interest. This design avoids hand-tuning the weights of different objectives and improves computational efficiency and optimization convergence.

\begin{figure}[t!]
\centering
\includegraphics[trim=0 0 0 10, clip,width=1\columnwidth]{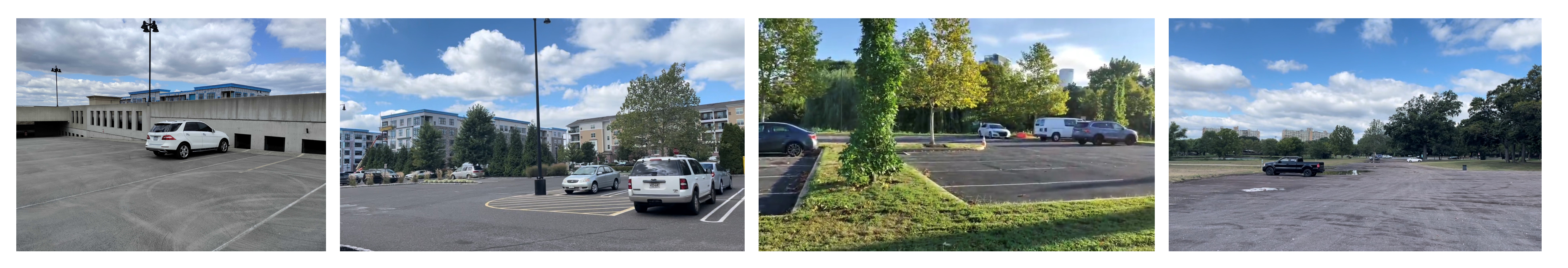}
    \caption{\textbf{Experiment environments} with vehicles, light poles, tree trunks. (Left) Garage. (Middle Two) Urban. (Right) Dirt road.}
    \label{fig:environments}
\end{figure}


\section{Proposed Approach}
\label{sec:proposed-approach}

\subsection{Object detection and modeling}
For each object geometry type, we construct a virtual sensor, which given raw point cloud data outputs the estimated object configuration measurements in the body frame. This pipeline has three steps: semantic segmentation, instance extraction, and object modeling. 

For semantic segmentation, we build upon RangeNet++~\cite{milioto2019rangenet++}, and drastically reduce the number of layers in the encoder and decoder to boost efficiency ($\sim$800\% the speed of the default DarkNet-53 \cite{redmon2018yolov3} backbone). It runs in real time on the Intel NUC computer.

The extraction and modeling of cylindrical objects and local ground planes are done in a similar way as presented in our previous work \cite{chen2020sloam}, and the root $\mathbf{b}$ of the cylinder model is the intersection of cylinder axis and local ground plane. For cylindrical objects, the resulting measurement is $\ell^{g}(\mathbf{z}) = [ \mathbf{b}{(\mathbf{z})}; \mathbf{n}{(\mathbf{z})}; r{(\mathbf{z})}]$, but in robot body frame.

For cuboidal objects, the resulting measurement is of the form $\ell^{g}(\mathbf{z})= [\mathbf{r}{(\mathbf{z})}; \mathbf{t}{(\mathbf{z})}; \mathbf{d}{(\mathbf{z})}]$, but again in robot body frame. Cuboidal objects are difficult to model from only one LIDAR scan. Therefore, we use LIDAR odometry to accumulate a 1$\sim$3 seconds of semantically segmented point clouds and filter points into a certain elevation window to reject outlier points. Filtered points with target class labels are clustered using DBSCAN~\cite{ester1996density-dbscan}. We then project the points within each cluster onto the ground plane, and extract the 2D convex hull of each cluster. We perform principal component analysis (PCA) on the 2D convex hull points to estimate the longitudinal axis $\mathbf{b}_1$ of the cuboid (first PCA component). We assume that the front of the vehicle is lower than the rear to determine the facing direction. 
 The vertical axis $\mathbf{b}_3$ of the cuboid is assumed to be the same as the normal of its nearby local ground plane. The lateral axis is then $\mathbf{b}_2 = \mathbf{b}_3 \times \mathbf{b}_1$. The dimensions $\mathbf{d}{(\mathbf{z})}$ = [$d_x$,  $d_y$,  $d_z$]$^\intercal$ are estimated by the distance between the 5th and 95th percentiles of projections onto the axes, and $\mathbf{b}_2$, $\mathbf{b}_3$ define the rotation $\mathbf{r}{(\mathbf{z})}$. The centers of the projections define the translation $\mathbf{t}{(\mathbf{z})}$.

\subsection{Metric-Semantic SLAM}
\label{sec:semantic-slam}


\begin{figure}[t!]
\centering
\includegraphics[trim=0 20 15 10, clip,width=1.0\columnwidth]{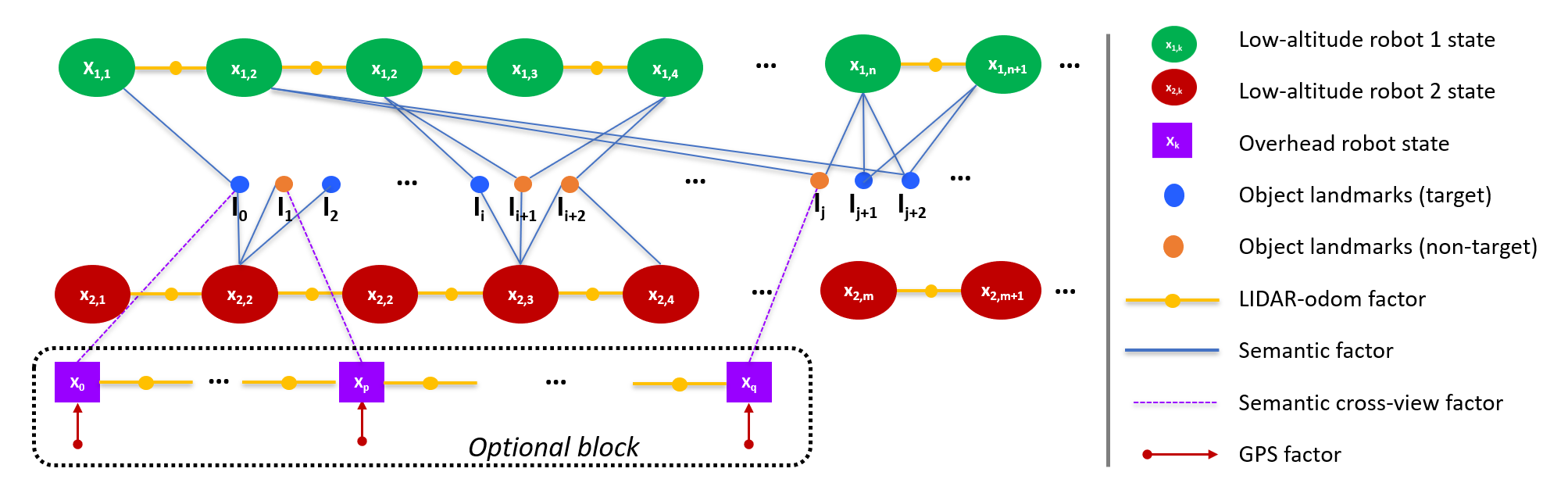}
\vspace{-0.1in}
    \caption{Metric-semantic SLAM factor graph representation. We define customized factors for cuboidal and cylindrical objects.}
    \label{fig:factor-graph}
\end{figure}



We build our semantic SLAM implementation on the GTSAM library \cite{gtsam,kaess2012isam2,kaess2008isam}. The factor graph of our semantic SLAM module is shown in \cref{fig:factor-graph}. The objects are classified into target and non-target objects, which are discussed in \cref{subsec:active metric-semantic mapping}. 



Let $\mathbf{H}^w_s$ be the matrix form of the robot pose $\mathbf{x}_t$,
where ${\mathbf{R}^w_s}$ is the rotation component and $\mathbf{t}^w_s$ is the translational component. We suppress the $i$ subscript of $\ell_i$ for notational compactness. Following \cite{gtsam,kaess2012isam2}, the measurement likelihood function with the Gaussian noise model is: $\mathbf{L}(\mathbf{x}, \ell^g; \mathbf{z}^g) = \mathbf{exp}\{-\frac{1}{2} ||\mathbf{h}(\mathbf{x}, \ell^g) \ominus \mathbf{z}^g||^2_\Sigma\}$, where $\mathbf{h}(\mathbf{x}, \ell^g) \ominus \mathbf{z}^g$ is the error $\mathbf{e_{(\cdot)}}$ that we will define separately for cuboid and cylinder objects, $\mathbf{z}^g$ is the output of the object detection and modeling step, and $\Sigma$ is the covariance matrix. 

\subsubsection{Odometry factor}
We use a LIDAR odometry algorithm \cite{qu2022llol} to generate the odometry factor. We calculate a relative $\mathbb{SE}$(3) transform between two consecutive pose estimates and use this transform as a factor between the corresponding poses in the graph. 

\subsubsection{Data association}
When detecting an object, we first associate it with objects in $\mathcal{M}$ or create a new map object. We employ nearest neighbor (NN) matching of the centroids (for cuboids) or roots (for cylinders), with a fixed threshold for valid matches, using the currently estimated robot pose to transform detected objects into the world frame for matching.

\subsubsection{Custom cuboid and cylinder factors} 
\label{sec:custom cuboid and cylinder factors}
Once we have object associations, we use these observation-landmark matches to form factors in the factor graph. To acquire the expected measurement of the cuboid, we can first transform the pose of the cuboid into the robot body frame ${\mathbf{H}^s_\texttt{cub}} = \mathbf{H}^s_w \mathbf{H}^{w}_\texttt{cub}$. We use a similar measurement error function as the one in~\cite{yang2019cubeslam} for cuboidal objects, i.e.,
\begin{equation}
\begin{aligned}
\mathbf{e}_\texttt{cub} = 
\begin{bmatrix}
\mathbf{log}((\mathbf{H}^{s}_\texttt{cub}(\mathbf{z}))^{-1} ({\mathbf{H}^s_w} \mathbf{H}^{w}_\texttt{cub}))^{\vee} \\ \mathbf{d} - \mathbf{d}{(\mathbf{z})}
\end{bmatrix}
\end{aligned}
\end{equation}
where $\vee$ is vee operator that maps the $\mathbb{SE}$(3) transformation matrix into  $6\times1$ vector, $\mathbf{log}$ is the log map, and $(\cdot){(\mathbf{z})}$ are the object measurements. Intuitively, this measurement error is the distance between the currently detected and the expected cuboid models in the tangent space of $\mathbb{SE}$(3) and the $3\times1$ dimension vector.

Similarly, we can calculate the expected measurement and actual measurement of cylinder objects from $\ell_i^{g}$ and $\ell_i^{g} (\mathbf{z})$. We define the measurement error function for cylindrical objects as:
\begin{equation}
\mathbf{e}_\texttt{cyl} 
= 
\begin{bmatrix}
(\mathbf{R}^s_w\mathbf{b} + \mathbf{t}^s_w) - \mathbf{b}{(\mathbf{z})}\\ 
\mathbf{R}^s_w\mathbf{n} - \mathbf{n}{(\mathbf{z})}\\ 
\mathbf{r} - \mathbf{r}{(\mathbf{z})}
\end{bmatrix}
\end{equation}

We implemented our custom factors to be compatible with the GTSAM library~\cite{gtsam}.

\subsubsection{Multi-robot Semantic SLAM}
Each robot maintains its own factor graph. When robots establish communication, they exchange their history object detections and odometry, with each other. Robots can treat other robots' data in the same way as their own by maintaining an odometry graph for each robot and performing data association in the same way as their own measurements. This merged factor graph is illustrated in \cref{fig:factor-graph}.

\subsection{Uncertainty modeling}
\label{sub-sec:uncertainty-modeling}

\begin{figure}[t!]
        \centering
        \begin{subfigure}[t]{0.185\columnwidth}
        \centering
        \includegraphics[trim=10 5 10 10, clip,width=1.1\textwidth]{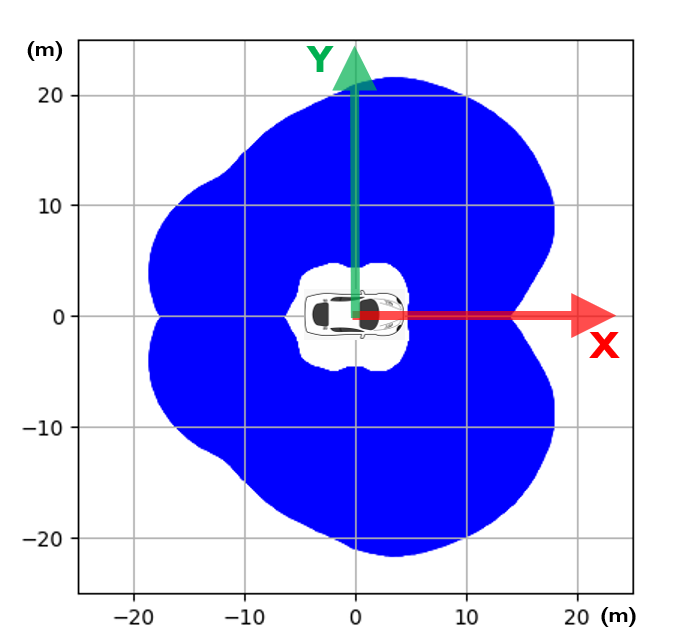}
        \end{subfigure}
        \begin{subfigure}[t]{0.185\columnwidth}
        \centering
        \includegraphics[trim=0 0 0 40, clip,width=1.1\textwidth]{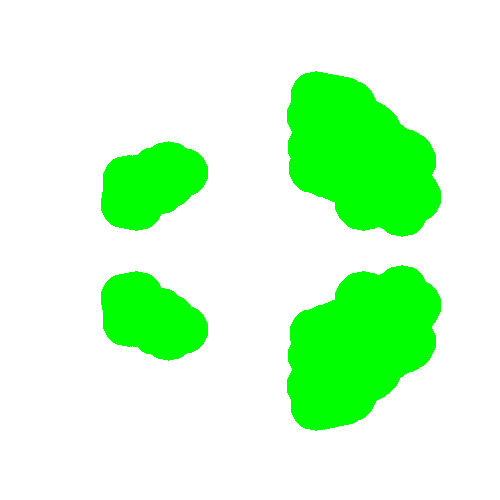}
        \end{subfigure}
        \begin{subfigure}[t]{0.185\columnwidth}
        \includegraphics[trim=0 0 0 40, clip,width=1.05\textwidth]{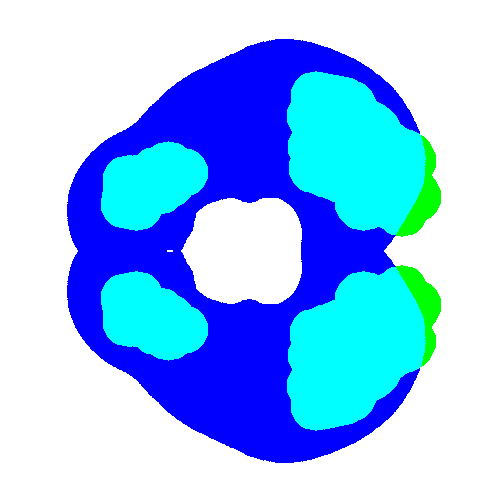}
        \end{subfigure}
        \begin{subfigure}[t]{0.185\columnwidth}
            \centering
            \includegraphics[trim=0 0 0 40, clip,width=1.05\columnwidth]{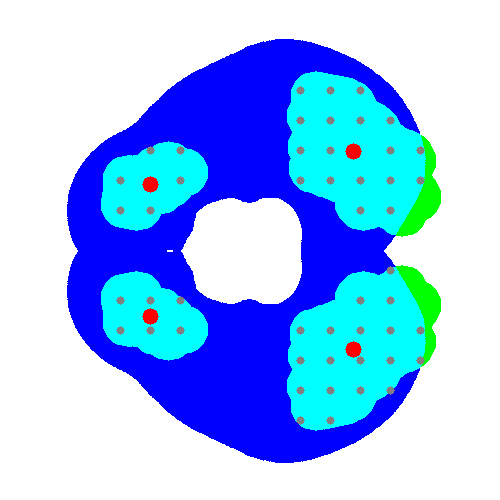}
        \end{subfigure}
        \begin{subfigure}[t]{0.185\columnwidth}
        \centering
        \includegraphics[trim=0 0 0 40, clip,width=1.05\columnwidth]{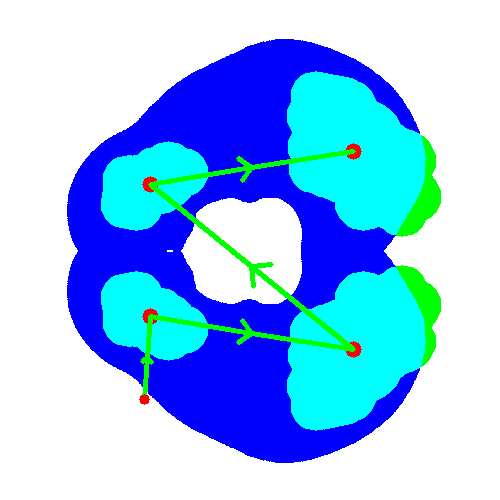}
            \end{subfigure}
        \caption{For a vehicle at (0,0) facing in the +x direction: (1st) Region with 95\% confidence in semantic classification.  (2nd) Region with 95\% confidence in localization and dimension estimation (with less than 0.1 m and 0.2 m error, respectively). (3rd) Regions with 95\% confidence level for both semantic and metric mapping are marked in cyan. (4th) Sampling. Two different sampling strategies: uniform sampling (grey) and centroid-only sampling (red), which form the set of valid samples, i.e., $\mathcal{X}^\texttt{visible}$. (5th) Best action sequence.}
        \label{fig:action-sampling}
        \label{fig:top-semantic-and-geoemtric-area-for-cars-all-figures-combined}
\end{figure}

As discussed in \cref{sec:introduction}, we seek to develop a pipeline for building models that map from the state space to the metric-semantic uncertainty space from real-world data. Formally, we want to find $f^s$ and $f^g$ such that $p_i = f^s(\mathbf{x}, \ell_i^g, \ell_i^s), 
\mathbf{\Sigma}(\ell_i^g) = f^g(\mathbf{x}, \ell_i^g, \ell_i^s)$, where $f^s$ maps state space into object classification confidence space, and $f^g$ maps state space into object geometric measurement uncertainty space.

\begin{figure*}[t!]
        \centering
    \includegraphics[trim=5 10 0 10, clip,width=1.0\textwidth]{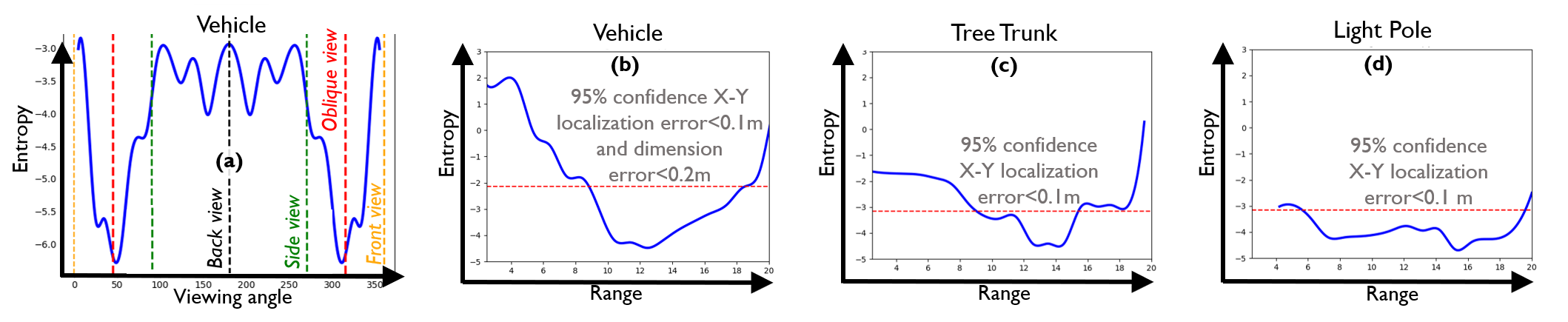}
        \caption{Geometric entropy ($\frac{1}{2}\ln\{ \det(\mathbf{\Sigma}(\ell^g_i)\}$) vs viewing angle (a). The oblique views lead to the least uncertainty in the vehicle geometric model. Geometric entropy vs range (b-d). For each semantic class, there is a range interval that leads to minimized geometric uncertainty.}
        \label{fig:angle-vs-car-geoemtric-error}
        \label{fig:top-region-for-geometric-error-for-car-lightpole-trunk}
\end{figure*}

\subsubsection{Acquiring the training samples}
To characterize the uncertainty, we propose extracting important low-dimensional inputs to reduce the dimensionality of the problem. From our data, we found that the most influential factors for object classification and modeling accuracies are range and viewing angles. Thus, we use these two quantities as inputs to the map predicting the uncertainty. To generate data, we manually fly a spiral-shaped trajectory centered around the object of interest to cover as many angles and ranges as possible. 

In order to model uncertainty, we require pairs of observation and ground truth to fit the model. To characterize the object classification uncertainty, we first extract the points enclosed by the cuboid or cylinder model estimated by our semantic mapping algorithm from the recent 10$\sim$30 time window. We then calculate the average per-point classification confidence from our semantic segmentation network, and use this as the object classification confidence. Once these samples are collected, we use a multilayer perceptron to approximate the underlying semantic uncertainty distribution and can then predict the confidence for any range and viewing angle, as shown in \cref{fig:top-semantic-and-geoemtric-area-for-cars-all-figures-combined}. A visualization is \href{https://youtu.be/S86SgXi54oU?t=100}{here}.

To obtain geometric model ground truth, we look up the actual dimensions of the vehicle online. To characterize geometric uncertainty, we first discretize the range and viewing angle. For each discretized interval, we calculate the error distribution of all measurements in the interval compared to the ground truth object model. A spline is fit to these data points which represent the underlying geometric uncertainties. The results of this characterization are shown in \cref{fig:top-region-for-geometric-error-for-car-lightpole-trunk}. Note that for vehicles we only consider uncertainties in X-Y localization, length, and width, since the Z position is well constrained by the ground, and height is observable from all viewing angles. For light poles and tree trunks, we only consider uncertainties in X-Y localization.

\subsection{PAMS}
\label{subsec:active metric-semantic mapping}
 The semantic classes are divided into target objects (vehicles), which are used to guide active semantic mapping, and non-target objects (tree trunks and light poles), which are only used to minimize robot localization drift.

\subsubsection{Target object discovery}
\label{subsubsec:high-level-planning}
With the object of interest locations extracted from the prior semantic map, we simply solve or approximate the Traveling Salesman Problem (TSP) to visit all of the objects as efficiently as possible. Our implementation also supports reactive exploration (active mapping upon detecting an object of interest), and it is trivial to incorporate a coverage planner into our system~\cite{liu2022large}.

\subsubsection{Uncertainty minimization}
\label{subsubsection:uncertainty minimization and action sampling}
As presented in \cref{sec:problem-formulation}, once the robot detects a target object it will minimize uncertainties in the geometric properties for the target object, while guaranteeing that the classification confidence and robot localization uncertainty are within a threshold. 

To guarantee object classification confidence, candidate actions are only sampled from the high confidence regions. To further reduce the sampling space, we also extract the low geometric uncertainty regions and samples in the intersection of the two, either by uniform sampling or by taking their centroids, as illustrated in \cref{fig:action-sampling}. 

We then generate candidate paths given the planning horizon $T$ constituting all possible orders of visiting the $a$ sample locations.  Therefore, the number of candidate actions is equal to the number of variations $P(a, T) = a\text{!} / (a-T)\text{!}$.

For each candidate path $\mathbf{x}_{\mathbf{\tau}} = [\mathbf{x}_{t+1}, \mathbf{x}_{t+1}, ... \mathbf{x}_{t+\tau}]$, we quantify the information gain. Supposing that $\ell^g_{i}$ is the target object to explore for time $t$, we use differential entropy to calculate the information gain as: $\mathbf{I}(\ell^g_{i}; \mathbf{z}_{\tau}) = \frac{1}{2}\ln\{ \det(\mathbf{\Sigma}(\ell^g_{i,t}))\} - \frac{1}{2}\ln\{\det(\mathbf{\Sigma}(\ell^g_{i,t+T}))\}$. We update the covariance recursively using the same method as shown in~\cite{atanasov2014information}: $\mathbf{\Sigma}(\ell^g_{i,t+1}) = (\mathbf{\Sigma}(\ell^g_{i,t})^{-1} + \mathbf{\Sigma}(\ell^g_{i,t+1})^{-1})^{-1}$, where $\mathbf{\Sigma}(\ell^g_{i,t+1})$ is estimated by the uncertainty models we obtained in \cref{sub-sec:uncertainty-modeling}. We choose the candidate path maximizing this information gain, and end exploration once entropy drops below a threshold or all samples are explored.

\section{Results and Analysis}
\label{sec:results and analysis}

\begin{figure*}[h!]
        \vspace{-0.05in}
        \centering
        \begin{subfigure}[t]{0.29\textwidth}
        \centering
        \includegraphics[trim=50 70 30 80, clip,width=0.95\columnwidth]{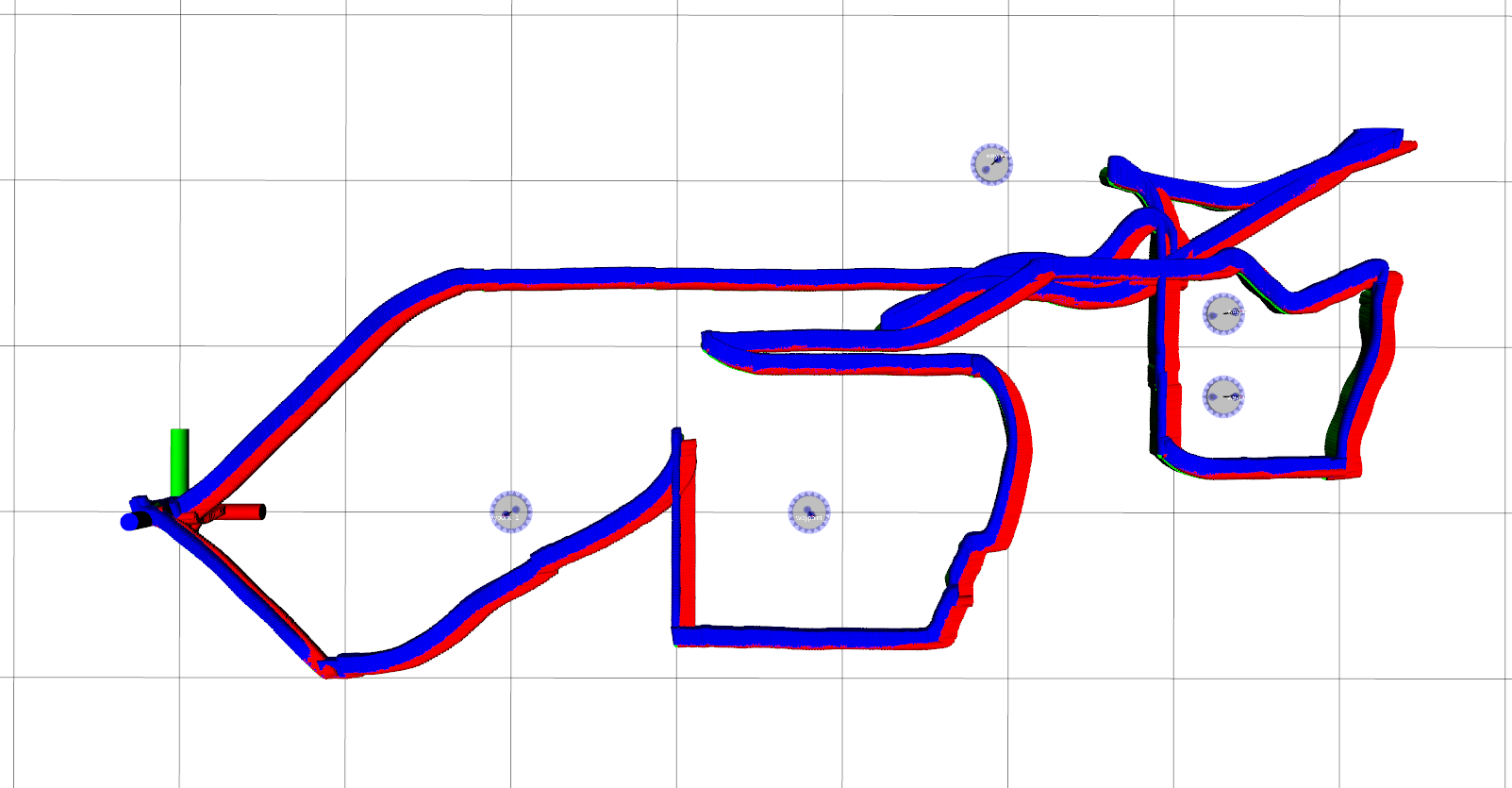}
        \end{subfigure}
        \begin{subfigure}[t]{0.29\textwidth}
        \centering
        \includegraphics[trim=50 70 30 80, clip,width=0.92\columnwidth]{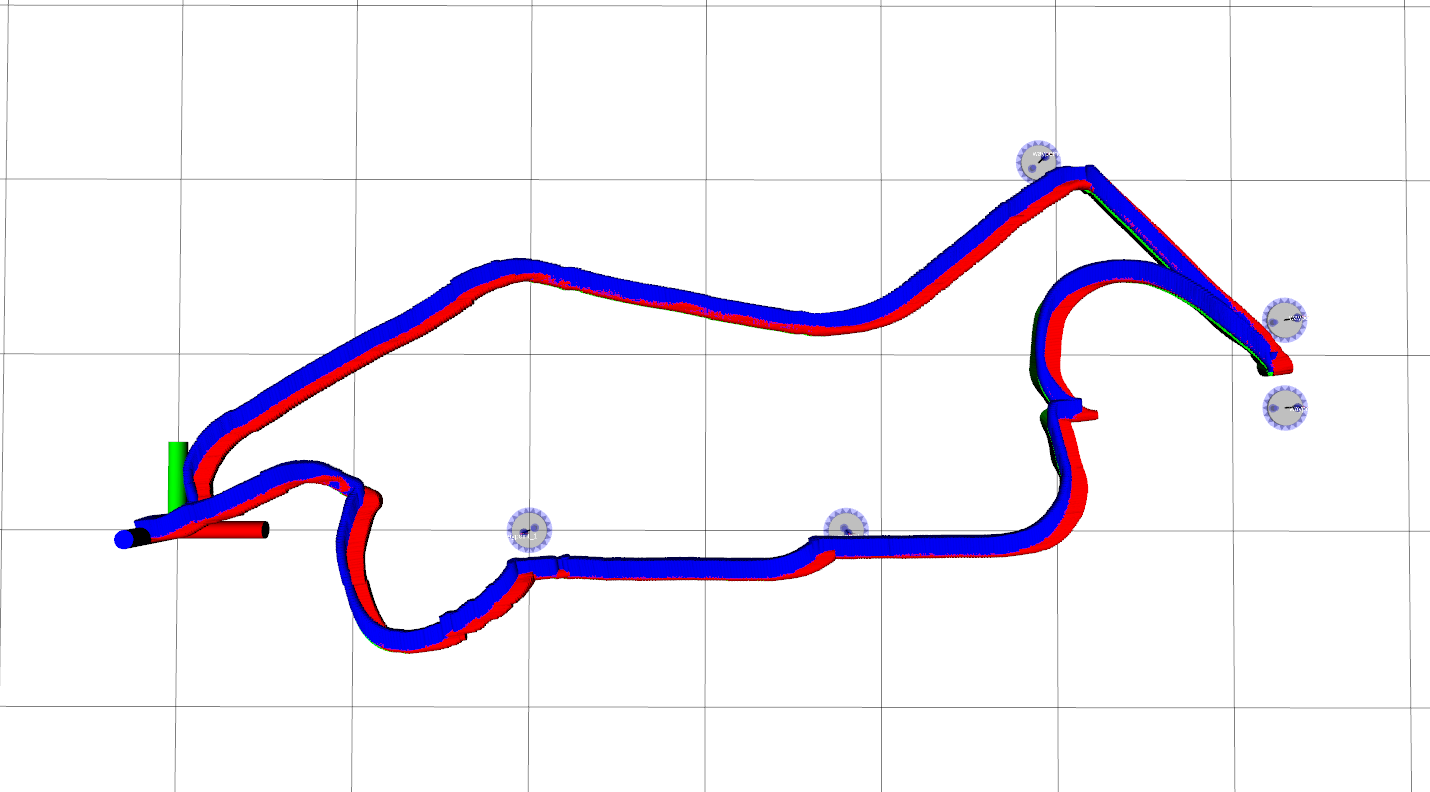}
        \end{subfigure}
        \centering
        ~~
                \begin{subfigure}[t]{0.18\textwidth}
        \includegraphics[trim=0 0 0 0, clip,width=0.8\columnwidth]{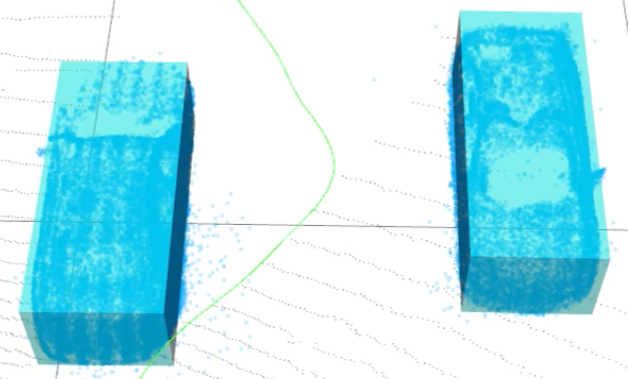}
        \end{subfigure}
        ~
        \begin{subfigure}[t]{0.18\textwidth}
        \centering
        \includegraphics[trim=0 0 0 0, clip,width=0.75\columnwidth]{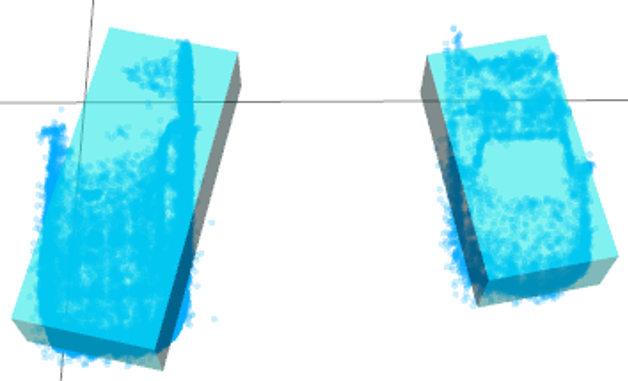}
        \end{subfigure}
        \caption{\textbf{Comparison of trajectory} examples from active metric-semantic mapping (leftmost) and heuristic-based mapping (middle left). Each grid is 10 m $\times$ 10 m. The ground-truth vehicle positions are marked by blue-colored disc-shaped markers. Our active metric-semantic mapping method drives the robot to observe objects from different ranges and viewing angles to minimize uncertainties. \textbf{Comparison of estimated vehicle model} examples from active metric-semantic mapping (middle right) and heuristic-based mapping (rightmost).}
        \label{fig:exploration-vs-coverage-instance-cuboid}
        \label{fig:exploration-vs-coverage-planning}
\end{figure*}

\begin{figure*}[t!]
\begin{center} 
 \setlength
 \resizebox{1.0\textwidth}{!}
 {\begin{tabular}{||c | c | c | c | c | c | c | c||} 
 \hline
 Measure & Urban Dataset 1 & Urban Dataset 2 & Urban Dataset 3 & Urban Dataset 4 & Parking Garage &  Dirt Road &  \textbf{All Data} \\
  \hline
  \hline
Err. Mean. (Ours / Baseline)  & \textbf{0.26 m} / 0.44 m &  \textbf{0.22 m} / 0.64 m &  \textbf{0.03 m} / 0.46 m &  \textbf{0.04 m} / 0.71 m &   \textbf{0.68 m} / 1.96 m  &  0.58 m / \textbf{0.39 m} &  \textbf{0.19 m} / 0.57 m \\
  \hline
 Err. Std. Dev. (Ours / Baseline) & \textbf{0.26 m} / 0.35 m &  \textbf{0.33 m} / 0.39 m &  \textbf{0.30 m} / 0.47 m &  0.36 m / \textbf{0.34 m} &  ---  &  0.20 m / \textbf{0.19 m} &  \textbf{0.39 m} / 0.55 m\\
    \hline
 Obj. Mapped (Ours / Baseline)  & {100}$\%$ /  {100}$\%$ & {100}$\%$ /  {100}$\%$ & {100}$\%$ / 100$\%$  & \textbf{100}$\%$ / {80}$\%$ & {100}$\%$ / 100$\%$ & {100}$\%$ / 100$\%$ & \textbf{100}$\%$ / {95}$\%$ \\
     \hline
\end{tabular}}
\end{center}
        \vspace{-0.1in}
\caption{Geometric model error mean and standard deviation and percentage of objects discovered.}
\label{fig:quant-result-table}
\end{figure*}

To demonstrate the performance of our system, we carried out experiments in various real-world environments as shown in \cref{fig:environments} which are diverse in object types, object shapes, and degree of structure. As illustrated in \cref{fig:front-page}, our complete system consists of both low-altitude and overhead UAVs, both built on our custom-made Falcon 4 UAV platform~\cite{liu2022large}. The low-altitude UAVs are equipped with forward-facing cameras and 3D LIDARs, and can autonomously navigate in cluttered and GPS-denied environments. To focus on the map quality, we assume that there is a prior map available (which could come from the high-altitude UAV) and the robots only explore the objects in this map. Our system detects and models vehicles (as cuboids), tree trunks and light poles (as cylinders), and the ground (as planes).

\subsection{Qualitative results}
\label{subsubsec:qual_results}

We observe in \cref{fig:top-semantic-and-geoemtric-area-for-cars-all-figures-combined} and \cref{fig:top-region-for-geometric-error-for-car-lightpole-trunk} that both semantic and geometric uncertainties vary significantly for the vehicle class when either the angle or the range changes, and these trends are clear and consistent across multiple datasets. However, for the tree trunk and light pole classes, the semantic confidence does not have a clear relationship with range or viewing angle. Even when there are very few points returned from the object, the segmentation network can classify them relatively well. This is probably because these objects are relatively distinct from the surrounding environment. There is a clear relationship, though, between the range and the geometric uncertainty for light poles and tree trunks, as seen in \cref{fig:top-region-for-geometric-error-for-car-lightpole-trunk}. Based on these observations, we model the distribution of metric and semantic uncertainties w.r.t. both range and viewing angle for the vehicle class, while only modeling the distribution of metric uncertainties w.r.t. range for the light pole and tree trunk classes.

As illustrated in \cref{fig:top-semantic-and-geoemtric-area-for-cars-all-figures-combined}, the classification confidence is higher from the two side views than from the front or back views, while the back view is more informative than the front view. This distribution is intuitive since from the side view, the vehicle is more readily identified. We also note from \cref{fig:top-semantic-and-geoemtric-area-for-cars-all-figures-combined} that the dimensions of vehicles are best estimated from oblique views. This is reasonable because, at oblique views, the sensor can observe both the lateral and longitudinal directions of the vehicle, resulting in a better fit.

In addition, our uncertainty characterization approach generalizes to different types of objects as in \cref{fig:top-region-for-geometric-error-for-car-lightpole-trunk}. The method can characterize uncertainties for any object of interest provided that the object shape can be approximated by cuboids or cylinders and the sensor used can output point cloud data. 

In \cref{fig:exploration-vs-coverage-planning} we compare the trajectories from active semantic mapping to our baseline method of visiting each object location in the prior map. While one could likely design a heuristic to perform similarly to our method, we emphasize that our system is rooted in real-world data instead of human intuition, and therefore is more generalizable. Our system achieves a 100\% detection accuracy and is significantly better in terms of object model accuracy than the heuristic-based method as shown on the right of the figure. The higher accuracy does come at the cost of a longer path length, but the difference in path length is less significant when the environment is more sparse because the robot will spend proportionally more time covering the environment as opposed to exploring individual objects.

It is obvious that the metric-semantic uncertainty distributions are highly non-linear and vary with object types, as shown in \cref{fig:top-region-for-geometric-error-for-car-lightpole-trunk}. It is difficult to design and verify mathematical models for such uncertainty distributions, especially for complicated objects that vary in type and appearance.

\subsection{Quantitative results}

We evaluate our system quantitatively using the accuracy of width and length estimates for target-class objects (i.e. vehicles) as the metric.  We looked up the dimensions based on the vehicle model to obtain the ground truth. For a discussion of localization error, we refer readers to our prior work \cite{chen2020sloam, liu2022large} which demonstrated the ability to reduce long-term drift in robot localization using semantic \gls{slam}.

We compare against our baseline method as described in \cref{subsubsec:qual_results}.
The object modeling accuracy is significantly improved using the proposed method, as shown in \cref{fig:quant-result-table}. The overall error mean is 0.19 m, which is around 3.5\% of the average vehicle dimensions. The overall error standard deviation is 0.39 m, which means that the vehicle dimensions can be estimated within 14.5\% error with more than 95\% confidence. By comparison, across all datasets, the error mean and standard deviation of the heuristic-based baseline method are around 300\% and 140\% of the error mean and standard deviation of our method. This shows that, by actively choosing the best viewing angles and ranges, the robot can gather significantly more accurate information.

We also note that there are edge cases where our method struggles. For example, the parking garage is a constrained space where most of the exploration viewpoints are unreachable. The dirt road is highly unstructured, lacking reliable landmarks such as tree trunks or light poles to constrain the robot's pose. Therefore, the longer the robot operates, the more drift accumulates, and the less accurate the model becomes. Thus, the heuristic-based approach that requires shorter flights outputs better results in this case. Additionally, Urban Dataset 1 contains all black vehicles, which are generally more difficult to detect, thus resulting in more noise. A potential solution is accounting for LIDAR intensities in our uncertainty characterization models.

\begin{figure}[h!]
\begin{center} 
 \setlength
 \resizebox{0.4\textwidth}{!}
 {\begin{tabular}{||c | c | c | c ||} 
 \hline
 Measure & Robot at 2.5 m & Robot at 5 m & Combined  \\
  \hline
   Num. of Obj. / Min  & 0.8 & \textbf{0.9} & 0.8  \\
 \hline
 Err. Mean  & 0.13 m & 0.15 m & \textbf{0.06 m}  \\
  \hline
 Err. Std. Dev. & \textbf{0.30 m} & 0.50 m & \textbf{0.30 m} \\
    \hline
\end{tabular}}
\end{center}
\vspace{-0.1in}
\caption{Mapping accuracy vs the number of heterogeneous robots.}
\label{fig:multi-robot-quant-result-table-multi-altitude}
\end{figure}


\begin{figure}[h!]
\begin{center} 
 \setlength
 \resizebox{0.485\textwidth}{!}
 {\begin{tabular}{||c | c | c | c | c ||} 
 \hline
 Measure & Robot in lot 1 & Robot in lot 2 & Robot in lot 3 & Combined \\
\hline
   Num. of Obj. / Min  & 0.8  & 1.6 & 1.6 & \textbf{3.3}  \\
 \hline
   Err. Mean  & 0.13 m & 0.22 m & \textbf{0.03 m} & 0.12 m \\
  \hline
 Err. Std. Dev. & \textbf{0.30 m} & 0.33 m & \textbf{0.30 m} & 0.33 m \\
    \hline
\end{tabular}}
\end{center}
\vspace{-0.10in}
\caption{Mapping efficiency vs the number of homogeneous robots.}
\label{fig:multi-robot-quant-result-table-coverage-area}
\end{figure}

We perform multi-robot experiments by flying sequentially and fusing the maps afterwards. Using heterogeneous robots leads to more accurate mapping results than any robot working independently as in \cref{fig:multi-robot-quant-result-table-multi-altitude}. Homogeneous robots can parallelize coverage linearly w.r.t. the number of robots without sacrificing map accuracy as in \cref{fig:multi-robot-quant-result-table-coverage-area}, assuming that the robots fly simultaneously.

Our full system runs in real time onboard UAVs with an Intel i7-10710U processor. Semantic segmentation runs at 1.5 Hz using only 1 out of 12 CPU threads. In total we use 5-6 threads, leaving $\sim$50\% computational headroom.

Finally, our map representation significantly reduces storage and communication burden. One million object models take only 9 MB of storage. By comparison, it requires 6,250 MB to store a 3D voxel map that covers the same area (0.1 m resolution, 20 m height, 5 m spacing between objects).

\section{Conclusion and Future Work}
\label{sec:conclusion}

Highly accurate metric-semantic maps are important for applications such as precision agriculture, infrastructure inspection, and asset mapping in factories. In this paper, we proposed an active metric-semantic mapping approach that enables robots to actively explore to minimize uncertainties in both metric and semantic information. We characterized the relationship between states and metric-semantic uncertainties empirically using a large amount of real-world data, and analyzed the resulting insights into the relationship between uncertainty and different viewpoints. Through real-world multi-robot experiments, we showed that our system is capable of constructing highly accurate metric-semantic maps in real time. We additionally showed that our active mapping system improved map quality, and using multiple heterogeneous robots can improve both the map accuracy as well as the mapping speed.  Future work includes scaling up to include more semantic classes, enabling robots to react to each other's information in real time, and considering more factors such as altitude angle and LIDAR point cloud intensity for the uncertainty characterization.


\bibliographystyle{IEEEtran}
\bibliography{ref, ag-survey-ref, phd_qual_ref}

\end{document}